\documentclass[a4paper,fleqn]{cas-dc}

\usepackage[numbers,sort&compress]{natbib}
\usepackage{amsmath,amssymb}
\usepackage{booktabs}
\usepackage{tabularx}
\usepackage{array}
\usepackage{makecell}
\usepackage{multirow}
\usepackage{microtype}
\usepackage{hyperref}

\hypersetup{
    hidelinks
}

\newcolumntype{Y}{>{\raggedright\arraybackslash}X}

\begin{document}

\let\WriteBookmarks\relax
\def\floatpagepagefraction{1}
\def\textpagefraction{.001}


\shorttitle{FedHisto-PAST for Lung Histopathology Classification}


\title[mode=title]{FedHisto-PAST: Parameter-Efficient Stain-Aware Federated Learning for Cross-Site Lung Histopathology Classification}


\shortauthors{M. M. Shahriar et al.}

\shorttitle{FedHisto-PAST for Lung Histopathology Classification}

\author[1]{Muhammad Muhtasim Shahriar}[orcid=0009-0008-4954-3811]
\ead{shahriarmuhtasim@gmail.com}

\author[2]{M.\,M. Golam Hafiz}[orcid=0009-0003-2795-1033]
\ead{rafihafiz2016@gmail.com}

\author[3]{Saad Aloteibi}
\ead{SaadAloteibi@ksu.edu.sa}

\author[4,5]{Mohammad Ali Moni}[orcid=0000-0003-0756-1006]
\cormark[1]
\ead{m.moni@uq.edu.au}

\affiliation[1]{
    organization={Department of Computer Science, International Islamic University Chittagong (IIUC)},
    city={Chittagong},
    country={Bangladesh}
}

\affiliation[2]{
    organization={Department of Computer Science, American International University-Bangladesh (AIUB)},
    city={Dhaka},
    country={Bangladesh}
}
\affiliation[3]{
    organization={Department of Computer Science and Engineering, College of Applied Studies, King Saud University},
    city={Riyadh},
    postcode={11437},
    country={Saudi Arabia}
}

\affiliation[4]{
    organization={AI \& Digital Health Technology, Artificial Intelligence and Cyber Futures Institute, Charles Sturt University},
    city={Bathurst},
    state={NSW},
    postcode={2795},
    country={Australia}
}

\affiliation[5]{
    organization={AI \& Digital Health Technology, Rural Health Research Institute, Charles Sturt University},
    city={Orange},
    state={NSW},
    postcode={2800},
    country={Australia}
}
\cortext[cor1]{Corresponding author}


\begin{abstract}
Cross-site lung histopathology classification must account for stain variation, non-IID client data, missing classes, and the cost of adapting large pathology encoders. This study evaluates FedHisto-PAST v2 for three-way classification of adenocarcinoma (ACA), Normal, and squamous cell carcinoma (SCC). FedHisto-PAST v2 combines a frozen HIBOU-B foundation model with parameter-efficient adaptation, stain-conditioned paired-view prediction and feature consistency, reliability-aware prototype learning, and adaptive federated aggregation. Experiments used a five-client, non-IID, raw-data-local simulation with fixed internal evaluation, client-level analysis, component ablations, communication accounting, and a development-influenced exploratory LungHist700 cohort. All principal methods achieved near-ceiling internal performance, which limited discrimination on the fixed split. On LungHist700, FedHisto-PAST v2 achieved a Macro-F1 of 0.728560 and a balanced accuracy of 0.730454. Higher recognition of Normal and SCC was accompanied by lower ACA recall, and calibration remained imperfect. Prediction-level consistency was the only component with a clearly supported independent contribution in the external ablation analysis. Feature consistency and prototype regularization showed no conclusive independent overall gains in Macro-F1. The framework updated 1.253841\% of the model parameters. The results provide exploratory cross-dataset evidence for stain-aware, parameter-efficient federation; they do not establish formal privacy, patient-level independence, prospective deployment, or clinical validation.
\end{abstract}



\maketitle


\section{Introduction}
\label{sec:introduction}

Histopathological examination remains central to distinguishing lung adenocarcinoma (ACA), squamous cell carcinoma (SCC), and non-malignant tissue because these categories exhibit different morphological patterns and carry different clinical interpretations. Automated analysis of digitized histology can support reproducible and scalable assessment of these patterns, and transfer-learning studies in lung and colon histopathology show that computational models can distinguish malignant from benign tissue categories in routine images \cite{ref01}. However, high internal accuracy on a curated image collection does not establish robustness across laboratories, staining workflows, scanners, or under-represented classes. The practical objective is to preserve morphology-sensitive discrimination when data are distributed across institutions and deployment conditions differ from those of the training environment.

Multi-institutional model development is constrained by practical, ethical, legal, and governance barriers to pooling patient-related pathology images. Decentralized swarm learning has therefore been used to train across partners while avoiding direct data transfer \cite{ref02}. Federated computational pathology similarly enables collaborative learning from distributed data silos, but its value depends on how communication, privacy mechanisms, and site heterogeneity are handled \cite{ref03}. A recent systematic review identified persistent challenges in communication overhead, computational demand, non-independent and identically distributed (non-IID) data, and inconsistent evaluation and reporting in federated histopathology classification \cite{ref04}. Histopathology is particularly susceptible to feature shifts because staining protocols, scanner configurations, and institutional artifacts vary across centers and can destabilize standard aggregation \cite{ref05}. Heterogeneous clients may also have long-tailed class counts or omit a class entirely, making global accuracy an incomplete measure and motivating class-balanced objectives that account for diminishing information from repeated majority-class samples \cite{ref06}. At the same time, full adaptation of large pathology foundation models would optimize and exchange a much larger trainable state than may be necessary for a specific task, creating a strong case for parameter-efficient adaptation within the federated loop.

Existing work addresses individual aspects of this problem, but their interaction remains unresolved. Orchestral stain-normalization GANs and federated stain-normalization systems show that laboratory-specific colour variation can be addressed without pooling raw slides \cite{ref07,ref08}. FedSDA and FedStain frame stain statistics as a source of non-IID feature shift, although their alignment mechanisms differ from paired counterfactual consistency in a frozen foundation-model adaptation pipeline \cite{ref09,ref10}. Pathology foundation models, such as HIBOU, provide reusable transformer representations, but their features can retain site-specific signatures that impair out-of-domain generalization and require task-aware adaptation \cite{ref11,ref12}. Low-Rank Adaptation (LoRA) reduces the trainable state of a frozen transformer through low-rank updates, providing a practical basis for federated parameter-efficient fine-tuning \cite{ref13}. Prototype-based federated learning offers a complementary mechanism for exchanging class-level representations across statistically and structurally heterogeneous clients \cite{ref14}. The remaining gap is a raw-data-local lung histopathology framework that jointly combines stain-conditioned parameter-efficient foundation-model adaptation, paired stain-counterfactual prediction and feature consistency, class-availability-aware reliable prototypes, and failure-aware multi-seed evaluation with explicit calibration, client robustness, statistical analysis, communication accounting, and privacy boundaries.

To address this gap, this paper proposes FedHisto-PAST v2, a parameter-efficient and stain-aware federated framework for three-class lung histopathology image classification. The framework is studied in a five-client non-IID federated simulation in which raw images remain local, without claiming differential privacy, secure aggregation, or any other formal privacy guarantee. A pretrained HIBOU-B encoder remains frozen, while adaptation is achieved via low-rank, lightweight trainable modules. Paired stain-counterfactual views preserve shared tissue geometry while encouraging consistent predictions and representations across colour perturbations. Class prototypes are constructed only from locally available classes and are weighted to limit the influence of unreliable client-class estimates. The design treats stain shift, label skew, and parameter cost as coupled problems: stain perturbations operate at the image level, consistency objectives operate at the prediction and representation levels, and prototype regularization transfers class-structured information only when local evidence exists. This coupling avoids forcing an absent class to be included in a client update while retaining a shared global decision space. The evaluation uses fixed internal partitions, client-level robustness, calibration, failure reporting, and a development-influenced exploratory cross-dataset cohort; it does not constitute prospective, patient-level, or clinical validation.

This paper makes five contributions. First, it formulates a federated architecture that combines a frozen HIBOU-B pathology foundation model with parameter-efficient trainable modules, limiting adaptation to a compact communicated state. Second, it introduces paired stain-counterfactual learning with prediction-level and feature-level consistency so that matched tissue content is compared under independently perturbed stain appearances. Third, it develops class-availability-aware and reliability-aware prototype regularization that excludes absent client classes and tempers unstable prototype contributions, without presuming that this component independently improves every endpoint. Fourth, it establishes a failure-aware evaluation protocol spanning multiple stochastic seeds, fixed internal testing, client robustness, calibration, the development-influenced exploratory LungHist700 cross-dataset cohort, and paired image-level bootstrap analysis. Fifth, it reports trainable parameters and communication accounting while maintaining explicit claim boundaries regarding raw-data locality, privacy, statistical independence, generalization, and clinical use. The integrated design and controlled evaluation constitute the contribution, with component-specific claims restricted to the reported ablation evidence.



\section{Related Work}
\label{sec:related_work}

\subsection{Computational Lung Histopathology Classification and Stain Variation}
\label{subsec:lung_histopathology_stain}

Patch-level lung histopathology studies show that transfer learning can distinguish adenocarcinoma, squamous cell carcinoma, and non-malignant tissue, but performance on a curated and augmented collection does not establish robustness to laboratory shift \cite{ref01}. A federated LC25000 study distributed the lung and colon classification task across clients, but its evidence remained tied to the same curated image source and conventional convolutional adaptation \cite{ref15}. These studies support computational discrimination of lung tissue categories and motivate evaluation beyond a single source, split, or staining environment.

Adjacent centralized histopathology research has compared binary wandmulticategory classification ,and magnification-specific with magnification-independent training \cite{ref16}. Collaborative transfer branches have been used to combine generic pretrained features with task-specific residual representations \cite{ref17}. Other work has integrated convolutional and transformer models for patch and whole-slide breast cancer subtype identification \cite{ref18}. Bidirectional recurrent fusion has been proposed to capture contextual relations in multiclass histology \cite{ref19}. For three-class lung classification, deep features have also been coupled with conventional machine-learning classifiers instead of using end-to-end training \cite{ref20}. These approaches broaden the available architectural choices, although most still rely on centralized image access and are evaluated within narrow source domains.

Transformer studies show that pretraining, colour normalization, augmentation, and patch configuration can materially affect histopathology performance and transferability \cite{ref21}. Multigranular fuzzy guidance addresses coarse, medium, and fine tissue structure, whereas attention-based feature fusion targets redundancy between scales \cite{ref22}. Specialized convolutional systems have also optimized binary and eight-class breast histology classification \cite{ref23}. Multi-resolution vision transformers extend this view to slide-level skin cancer classification across several magnifications and an external cohort \cite{ref24}. Multimodal fusion studies combine histology with clinical variables, showing that tissue appearance is only one component of some diagnostic tasks \cite{ref25}. At cellular resolution, StarDist-based work shows that stain augmentation, imbalance handling, and ensembling remain important even when the target is nuclei segmentation and classification \cite{ref26}. Hybrid EfficientNetV2-transformer models further demonstrate high-capacity feature fusion, although the evidence reported is limited to a single breast histology dataset \cite{ref27}.

More broadly, laboratory protocols, scanners, tissue processing, and staining chemistry can create site-linked appearance cues that centralized models may exploit. Colour normalization can reduce visible chromatic differences, but it does not necessarily remove complex site signatures from learned pathology representations \cite{ref12,ref21}. Cross-site lung histopathology classification, therefore, requires methods that separate disease morphology from stain-dependent appearance while preserving minority-class evidence.

\subsection{Federated Learning in Computational Pathology}
\label{subsec:federated_computational_pathology}

Federated averaging established the basic protocol of repeated local optimization followed by server-side weighted model aggregation \cite{ref28}. FedProx generalized this protocol by adding a proximal objective that limits local drift under statistical and systems heterogeneity \cite{ref29}. FedBN keeps batch-normalization parameters local to address feature-distribution shift between clients \cite{ref30}. SiloBN applied the same local-statistics principle to multicentric histopathology and reported improved transfer across centres \cite{ref05}. These methods show that non-IID pathology includes both label skew and client-specific appearance shift, either of which can distort representations and aggregation.

Histopathology-specific federated learning has progressed from weakly supervised whole-slide learning with optional differential privacy to task-specific strategies for classification, segmentation, and deployment \cite{ref03}. FedSODA uses synthetic cross-assessment and dynamic layer-wise aggregation to address sample imbalance and organ-dependent heterogeneity in federated segmentation \cite{ref31}. Federated magnification fusion combines information from multiple resolutions for breast tumour classification, but it does not explicitly model stain-counterfactual invariance \cite{ref32}. FedDP adds differential-privacy noise to model updates for histopathology segmentation, providing a formal privacy mechanism beyond raw-data locality \cite{ref33}. An end-edge nuclei framework combines FedAvg with model quantization, emphasizing deployment efficiency rather than cross-site stain adaptation \cite{ref34}. FedWSIDD exchanges distilled synthetic slides, enabling clients to retain heterogeneous multiple-instance learning architectures \cite{ref35}.

Stain-aware federation follows several routes. Orchestral normalization aligns client appearance through generative translation \cite{ref07}, while BottleGAN learns federated stain mappings for downstream weight aggregation \cite{ref08}. FedSDA aligns stain distributions using stain separation and a diffusion model trained on stain matrices rather than raw images \cite{ref09}. FedStain models higher-order stain statistics for federated domain generalization, focusing on distributional descriptors rather than paired views of the same tissue \cite{ref10}. These approaches treat stain variation as a federated feature-shift problem, but their normalization, synthesis, or distribution-alignment objectives differ from those that enforce agreement between predictions and representations across matched counterfactual views. Retaining raw images locally reduces direct data movement, but it does not provide differential privacy, secure aggregation, or protection against update-level attacks unless those mechanisms are explicitly implemented \cite{ref03,ref33}.

\subsection{Pathology Foundation Models and Parameter-Efficient Adaptation}
\label{subsec:pathology_foundation_models}

Pathology foundation models seek transferable representations by pretraining vision encoders on large, heterogeneous collections of slides. HIBOU provides pathology-specific vision transformers pretrained on self-supervised data across diverse tissues and stains \cite{ref11}. Virchow showed that large-scale pathology pretraining can support clinical-grade and rare-cancer tasks through transferable image representations \cite{ref36}. TITAN extends this direction with multimodal whole-slide pretraining that connects visual and language information \cite{ref37}. Prov-GigaPath learns whole-slide representations from real-world pathology data at scale \cite{ref38}. Nevertheless, independent benchmarking shows that the usefulness of a frozen foundation-model feature extractor depends on the downstream task, aggregation method, and evaluation protocol \cite{ref39}.

Frozen extraction reduces training cost and may stabilize small-data learning, but a fixed representation cannot automatically remove institution-specific shortcuts. FLEX shows that pathology foundation-model features can encode site-specific and demographic signals that impair performance on unseen domains \cite{ref12}. An adjacent ECG foundation-model study demonstrates the broader value of external and zero-shot evaluation, although its signal modality and diagnostic setting are not directly comparable to histopathology \cite{ref40}. Task-specific vision transformer systems such as HistoViT also achieve broad histopathology classification performance, but they require conventional task adaptation rather than federated parameter-efficient tuning \cite{ref41}.

Low-rank adaptation freezes pretrained weights and learns compact rank-decomposition updates, reducing the trainable state required for downstream adaptation \cite{ref13}. This property is relevant in federation because only lightweight task-specific modules need to be optimized, transmitted, and stored. However, parameter efficiency alone does not address stain shift, missing client classes, unreliable local statistics, or conflicting updates. A pathology-oriented federated design must therefore connect lightweight adaptation with stain-aware objectives and aggregation rules that remain meaningful under client heterogeneity.

\subsection{Consistency Learning, Prototype Methods and Reliability Evaluation}
\label{subsec:consistency_prototypes_evaluation}

Consistency regularization provides a direct mechanism for ensuring that matched samples retain similar predictions or representations under controlled perturbations. Existing stain-aware federated work primarily aligns generated styles, stain distributions, or higher-order statistics rather than jointly constraining paired prediction and feature agreement \cite{ref07,ref10}. A paired stain-counterfactual formulation treats two stain-conditioned views of the same tissue as semantically equivalent, allowing the learning objective to penalize stain-sensitive disagreement. This formulation complements distribution-level alignment because it operates on sample correspondence rather than only population statistics.

Prototype-based federation transfers class-level feature summaries and can reduce dependence on gradient alignment across heterogeneous clients \cite{ref14}. Because a class prototype is computed from observed local examples, a client with no samples from a class cannot provide a valid estimate for that class \cite{ref14}. Severe class imbalance further makes prototype quality unequal even when a class is present, motivating class-availability checks and reliability weighting. Class-balanced objectives based on the effective number of samples provide one complementary response to unequal class frequency \cite{ref06}. Reliability-aware prototypes can therefore be interpreted as semantic regularizers whose contribution depends on the volume and stability of evidence, rather than as uniformly trustworthy global targets.

Evaluation practice is equally important because near-ceiling aggregate accuracy can conceal client-specific failures, class-specific sensitivity loss, or probability miscalibration. The recent systematic review of federated histopathology identified inconsistent reporting, communication burden, robustness, and multi-institutional validation as continuing concerns \cite{ref04}. Foundation-model benchmarks and external-domain studies likewise show that conclusions can change with task, site, cohort, and aggregation protocol \cite{ref12,ref39}. A defensible federated pathology evaluation should report multiple seeds, client-level and worst-client behaviour, calibration, communication cost, and external-domain performance. When formal privacy is absent, evaluation and reporting must also distinguish raw-data-local training from quantified privacy protection \cite{ref33}.

\subsection{Synthesis and Research Gap}
\label{subsec:synthesis_research_gap}

The literature has separately advanced non-IID optimization, local normalization, stain alignment, foundation-model pretraining, low-rank adaptation, and prototype exchange. Their integration into a single auditable lung-histopathology framework remains unresolved, especially when feature shift, label imbalance, missing classes, parameter transmission, and evaluation reliability must be addressed together \cite{ref04,ref09}. Existing foundation-model adaptation also shows that large pretrained encoders can preserve site-specific shortcuts, so freezing or fine-tuning an encoder is not sufficient for cross-domain robustness \cite{ref12}. The specific gap is the absence of a unified framework that combines frozen pathology-foundation-model parameter-efficient adaptation, client-side stain conditioning, paired stain-counterfactual prediction and feature consistency, missing-class- and reliability-aware prototypes, adaptive aggregation, multi-seed internal and client-level evaluation, calibration assessment, communication accounting, and exploratory cross-dataset evaluation.

FedHisto-PAST v2 is positioned to study this combined gap in a five-client non-IID simulation for ACA, Normal, and SCC classification. It uses a frozen HIBOU-B encoder with lightweight trainable adaptation, client stain descriptors, paired stain-counterfactual learning, class-availability- and reliability-aware prototype regularization, and adaptive aggregation. Its design is raw-data-local and privacy-aware, but provides no formal privacy guarantee. Its evaluation combines fixed internal testing with client robustness, calibration, and communication reporting, as well as a development-influenced exploratory LungHist700 cross-dataset cohort. The integrated methodology defines the question evaluated in this study; individual components are assessed only through the reported ablations.

\begin{table*}[t]
\centering
\caption{Positioning of closely related federated computational-pathology methods and FedHisto-PAST v2}
\label{tab:method_positioning}
\scriptsize
\setlength{\tabcolsep}{2pt}
\renewcommand{\arraystretch}{1.15}

\begin{tabularx}{\textwidth}{
@{}
>{\raggedright\arraybackslash}p{1.35cm}
>{\raggedright\arraybackslash}p{1.65cm}
>{\raggedright\arraybackslash}p{1.90cm}
>{\raggedright\arraybackslash}p{1.55cm}
>{\raggedright\arraybackslash}p{1.45cm}
>{\raggedright\arraybackslash}p{2.05cm}
>{\raggedright\arraybackslash}p{1.65cm}
>{\raggedright\arraybackslash}p{1.75cm}
Y
@{}
}
\toprule
Study &
Primary pathology task &
Federated/non-IID strategy &
Stain handling &
Foundation model or PEFT &
Prototype/consistency mechanism &
Formal privacy mechanism evaluated? &
External or unseen-domain evaluation &
Remaining gap relative to FedHisto-PAST v2 \\
\midrule

SiloBN \cite{ref05} &
Multicentric tumour-patch classification &
Centre-specific batch-normalization statistics with shared remaining parameters &
Local normalization statistics &
Not reported &
Not reported &
No formal mechanism evaluated &
Cross-dataset transfer between Camelyon cohorts &
No pathology foundation-model PEFT, paired stain consistency, or reliability-aware prototypes \\

Orchestral stain-normalization GAN \cite{ref07} &
Histopathology image analysis under distributed staining styles &
Federated learning with orchestral generative stain normalization &
GAN-based stain normalization &
Not reported &
No class-prototype mechanism reported &
No formal mechanism evaluated &
Multi-source evaluation reported &
Focuses on normalization rather than paired feature/prediction consistency and missing-class-aware prototypes \\

BottleGAN \cite{ref08} &
Federated computational-pathology segmentation &
Federated unsupervised stain mapping combined with weight aggregation &
Many-one-many generative stain normalization &
Not reported &
No prototype or paired-consistency mechanism reported &
No formal mechanism evaluated &
Heterogeneous multi-institutional evaluation &
Requires a generative normalization stage and does not address foundation-model PEFT or missing classes \\

FedSDA \cite{ref09} &
Non-IID histopathology classification &
Federated stain-distribution alignment using stain matrices and diffusion modelling &
Stain separation and distribution alignment &
Not reported &
No class-prototype mechanism reported &
Privacy-risk mitigation discussed; no formal guarantee evaluated &
Cross-client and non-IID evaluation reported &
Distribution alignment is not paired counterfactual consistency and does not federate lightweight foundation-model adapters \\

FedWSIDD \cite{ref35} &
Weakly supervised whole-slide classification &
Clients exchange distilled synthetic slides and may use heterogeneous MIL models &
Stain normalization inside dataset distillation &
Pretrained feature extraction; no federated PEFT reported &
No class-prototype or paired-consistency mechanism reported &
No formal mechanism evaluated &
Camelyon16 and Camelyon17 multi-centre evaluation &
Targets model heterogeneity through synthetic data exchange rather than stain-conditioned PEFT and reliability-aware semantic regularization \\

FedHisto-PAST v2 &
Three-class ACA/Normal/SCC image classification &
Five-client non-IID simulation with adaptive reliability-aware aggregation &
Client stain descriptors and paired stain-counterfactual views &
Frozen HIBOU-B with lightweight trainable adaptation &
Prediction/feature consistency and class-availability/reliability-aware prototypes &
No formal mechanism; raw-data-local simulation &
Fixed internal evaluation and development-influenced exploratory LungHist700 evaluation &
Integrated target framework; prospective clinical validation and formal privacy guarantees remain outside scope \\

\bottomrule
\end{tabularx}
\end{table*}

\section{Methodology}
\label{sec:methodology}

\subsection{Study Design and Reporting Scope}
\label{subsec:study-design}

This study used an image-level, three-class design to evaluate parameter-efficient federated learning for lung histopathology under controlled non-independent and identically distributed (non-IID) conditions. The target categories were adenocarcinoma (ACA), histologically normal lung tissue (Normal), and squamous cell carcinoma (SCC). Five simulated clients, C1--C5, participated in each communication round. These clients were dataset partitions rather than operational hospitals; therefore, the study is described as a five-client non-IID raw-data-local federated simulation. Source images and labels remained within their assigned partitions, while a shared frozen pathology encoder and lightweight trainable modules were coordinated through the federated loop.

The internal protocol used a fixed global validation partition, a fixed global test partition, and deterministic client-specific holdouts. Global-validation Macro-F1 governed checkpoint selection and early stopping. The global test set was evaluated only after the best checkpoint had been selected, and it was excluded from local optimization, server aggregation, hyperparameter choice, threshold selection, and stopping decisions. The design was neither leave-one-client-out nor cross-validation, so it did not evaluate generalization to unseen clients.

\begin{figure*}[t]
\centering
\includegraphics[width=\textwidth]{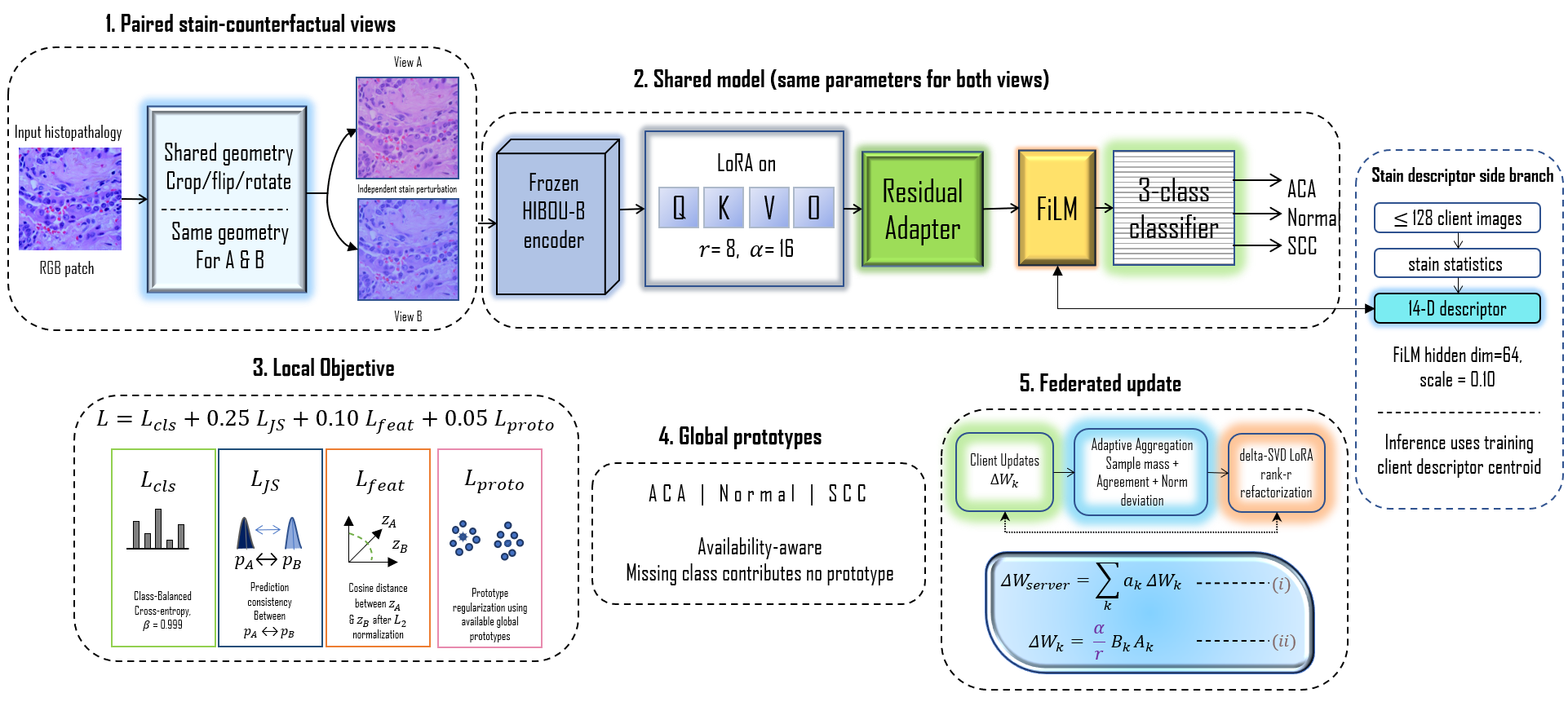}
\caption{Integrated FedHisto-PAST study and learning framework. (A) Canonical data clearance, C1--C5 non-IID partitions, fixed global validation and test sets, client holdouts, and the internal, ablation, and exploratory evidence tiers. (B) Shared-geometry stain-counterfactual views processed by frozen HIBOU-B, LoRA, residual adaptation, stain-conditioned FiLM, and supervised, prediction-consistency, feature-consistency, and prototype losses. (C) Local optimization, trainable-state transfer, class-availability masks, reliability-aware prototype exchange, adaptive model aggregation, and delta-SVD LoRA refactorization. (D) Validation-based checkpoint selection, fixed internal and client-level testing, five-checkpoint LungHist700 ensembling, and image-level paired bootstrap analysis without patient-clustered inference.}
\label{fig:fedhisto-framework}
\end{figure*}

Three evidence tiers were prespecified. The principal internal comparison covered FedAvg-PEFT, FedProx-PEFT, FiLM-Zero, FedHisto-PAST v1, and the final FedHisto-PAST configuration. Three component-removal experiments omitted prediction consistency, feature consistency, or prototype regularization, respectively. The frozen, validation-selected checkpoints were then assessed on LungHist700, which had influenced method development and was therefore treated as a development-influenced exploratory cross-dataset cohort rather than as an untouched external validation set. Figure~\ref{fig:fedhisto-framework} summarizes the integrated study, local learning, communication, and evaluation workflow. Formal privacy mechanisms were outside the executed protocol and are specified in Section~\ref{subsec:privacy-boundaries}.

\subsection{Dataset Construction and Client Partitioning}
\label{subsec:dataset-construction}

The canonical internal corpus contained 8,922 labelled images. Of these, 6,083 formed the five pre-holdout client pools, 1,419 formed the fixed global-validation partition, and 1,420 formed the fixed global-test partition. Deterministic class-wise extraction from the client pools yielded 5,325 optimization images and 758 client-specific holdout images. The public-source corpus incorporated lung histopathology material associated with LC25000 \cite{ref42}, as well as an external-source client derived from WSSS4LUAD resources \cite{ref43}. During reconstruction, duplicate LC25000/Kaggle mirror lineages were treated as a single source family. Native source dimensions and a complete patient-, slide-, and institution-level annotation chain were unavailable; therefore, the analytical unit was the labelled image.

The client design introduced both quantity and label skew. C1 was comparatively small and balanced; C2 contained SCC as a strong minority; C3 had greater SCC representation; C4 was SCC-majority; and C5 contained only ACA and Normal. Before holdout extraction, C5 included 1,181 ACA and 1,831 Normal images, and no SCC images. This missing-class condition was preserved in its optimization and holdout subsets, so no SCC example was created or reassigned to C5. C1--C4 used a 10\% local holdout, whereas C5 used 15\%. Within each available class, a seeded deterministic shuffle was used to select holdout images while retaining at least one optimization example whenever a class contained two or more images. The canonical split seed was 2026. For repeatable source-local evaluation, each run combined its stochastic seed with a fixed client-position offset while leaving the global validation and test assignments unchanged.

The global-validation partition contained 472 ACA, 474 Normal, and 473 SCC images; the global-test partition contained 473 ACA, 475 Normal, and 472 SCC images. These approximately balanced global partitions were fixed across valid methods and stochastic seeds. Portable relative-path hashes were used to generate stable sample identifiers for manifest control, prediction alignment, and paired statistical comparisons. These identifiers did not recover patient or whole-slide grouping. Table~\ref{tab:compact-partitions} reports the compact composition used throughout the canonical experiments.

\begin{table*}[t]
\centering
\caption{Compact client and partition composition}
\label{tab:compact-partitions}
\scriptsize
\setlength{\tabcolsep}{4pt}
\renewcommand{\arraystretch}{1.12}

\begin{tabularx}{\textwidth}{
@{}
>{\raggedright\arraybackslash}p{2.0cm}
>{\centering\arraybackslash}p{3.1cm}
>{\centering\arraybackslash}p{3.1cm}
>{\centering\arraybackslash}p{2.8cm}
Y
@{}
}
\toprule
Partition/client &
\makecell{Pre-holdout\\ACA/Normal/SCC} &
\makecell{Optimization\\ACA/Normal/SCC} &
\makecell{Holdout\\ACA/Normal/SCC} &
Principal non-IID characteristic \\
\midrule

C1 &
229/208/293 &
206/187/264 &
23/21/29 &
Relatively balanced; small \\

C2 &
485/439/65 &
437/395/59 &
48/44/6 &
SCC minority \\

C3 &
187/172/318 &
168/155/286 &
19/17/32 &
SCC enriched \\

C4 &
64/58/553 &
58/52/498 &
6/6/55 &
SCC majority \\

C5 &
1181/1831/0 &
1004/1556/0 &
177/275/0 &
External-source; SCC absent \\

Global validation &
472/474/473 &
N/A &
N/A &
Fixed; approximately balanced \\

Global test &
473/475/472 &
N/A &
N/A &
Fixed; approximately balanced \\

\bottomrule
\end{tabularx}

\vspace{2pt}
\begin{minipage}{\textwidth}
\footnotesize
\textit{Note.} Counts are ordered ACA/Normal/SCC. C1--C4 used 10\% local holdouts and C5 used 15\%. Global partitions were not subdivided into client optimization and holdout sets.
\end{minipage}
\end{table*}

\subsection{Dataset Clearance, Leakage Control, and Split Integrity}
\label{subsec:dataset-clearance}

Dataset clearance preceded model training. Every named image was opened and converted to RGB; any unreadable file raised an error rather than being replaced. Resolved paths were compared across client optimization sets, client holdouts, global validation, and global test. SHA-256 digests were then used to identify byte-identical content, and the strict execution criterion required zero exact cross-boundary matches. The canonical manifest binds each valid run to the same image inventory and partition assignments.

Two similarity-oriented audits supplemented exact hashing. A perceptual-family stage searched for probable transformed relatives, and a local-feature stage assessed image relationships using localized visual evidence. Under the implemented thresholds, the retained audit records contained zero cross-boundary family overlaps, including no detected global-validation/test family and no detected cross-client training family. Detailed matching fields, thresholds, and file-level evidence are retained for the Supplementary Methods rather than repeated in the main manuscript. The family-safe rebuild selected the retained representatives before final partition generation, after which the audit bundle recorded zero automatic family edges and zero unresolved scoring failures across the protected boundaries.

These checks establish an image-corpus quality-control boundary. They do not prove complete freedom from leakage. Exact hashes cannot detect every transformed copy, and similarity procedures remain sensitive to cropping, compression, stain alteration, and threshold choice. In addition, the unavailability of patient and slide identifiers prevented partitioning by patient or slide. The internal split is therefore described as path-disjoint, exact-file-disjoint, and audit-family-separated under the implemented procedures, without claiming biological or clinical independence.

\subsection{Image Preprocessing and Stain-Counterfactual View Generation}
\label{subsec:image-preprocessing}

Images were loaded in three-channel RGB format. For the final FedHistoa-PAST training, one random geometric transformation was sampled for each source image before creating the paired views. The shared transformation used a random resized crop to $224 \times 224$ pixels, a scale range of 0.85--1.00, and an aspect-ratio range of 0.90--1.10. Horizontal and vertical flips were applied with probabilities 0.5 and 0.2, respectively, followed by a uniformly sampled rotation between $-10$ degrees and $+10$ degrees with bilinear interpolation and white filling. Applying the same geometry to both branches ensured that consistency was evaluated across different stain realizations of the same tissue content rather than across unrelated crops.

The shared-geometry image was copied into views A and B, which were perturbed independently in optical-density space. RGB intensities were mapped to optical density, and a $3 \times 3$ transformation initialized to the identity was perturbed with additive Gaussian matrix noise with standard deviation 0.025. Its diagonal elements were independently scaled within 0.78--1.22. Optical-density values were clipped to 0--5 before conversion back to RGB. Each view then received independent brightness and contrast factors sampled over the range 0.92--1.08. These operations were designed to modify color and stain expression while preserving morphology; they were not considered to be pathologist-validated stain normalization.

After view generation, tensors were standardized with channel means of 0.485, 0.456, and 0.406 and standard deviations of 0.229, 0.224, and 0.225. Global validation, global testing, client-holdout evaluation, and LungHist700 inference used deterministic whole-image resizing to $224 \times 224$ followed by the same normalization. Baselines and FedHisto-PAST v1 used the canonical single-view augmentation pathway; the final method and its component ablations used the shared-geometry paired-stain pathway.

\subsection{Stain Descriptor and Stain-Conditioned FiLM}
\label{subsec:stain-descriptor}

A client-level stain descriptor supplied the conditioning signal for feature-wise linear modulation (FiLM). Descriptor estimation sampled at most 128 images per client using a deterministic, class-aware procedure and resized each image to $128 \times 128$ pixels. Tissue optical-density observations were used to estimate two ordered stain directions. Pixels with any optical-density channel below 0.15 were treated as background, and descriptor estimation proceeded when at least 50 tissue pixels remained. The two directions were deterministically ordered and normalized before calculating concentration summaries. The resulting 14-dimensional vector contained six direction elements and, for each estimated stain, the mean, standard deviation, first quartile, and third quartile of the concentration distribution. Descriptors were standardized across the participating training clients.

The shared FiLM network mapped the descriptor through a hidden layer with 64 units to produce multiplicative and additive feature modulations. The modulation scale was fixed at 0.10, limiting the magnitude of client-conditioned changes to the shared adapted representation. Client-specific holdout evaluation used the descriptor of the corresponding client. Global validation, global testing, and LungHist700 inference used a fixed centroid computed only from the five training-client descriptors, avoiding evaluation-derived conditioning.

FiLM-Zero retained the same trainable FiLM capacity but replaced every descriptor with the zero vector. It therefore controlled for additional network capacity without supplying client station information. The transmitted 14-dimensional summaries were not images, but their exchange was not assigned a formal privacy guarantee. Lower-level eigendecomposition, concentration-fitting, and fallback-descriptor procedures are reserved for the Supplementary Methods.

\subsection{Model Architecture and Parameter-Efficient Adaptation}
\label{subsec:model-architecture}

The shared encoder was HIBOU-B (\path{histai/hibou-b}), a pathology foundation model \cite{ref11}, loaded from repository revision \path{7fdc8cce365ebfd9e7e60ee6ee9a8d4d59388f3c} with Transformers 4.44.2 and the corresponding image processor. Pretrained initialization was mandatory, random initialization was disabled, and all backbone parameters remained frozen. Low-Rank Adaptation (LoRA) \cite{ref13} was attached to the transformer attention query, key, value, and attention-output dense projections. LoRA used rank $r=8$, scaling $\alpha=16$, and dropout 0.05. A residual adapter with a hidden dimension of 256 and dropout of 0.10, the three-class classifier, and FiLM, when enabled, formed the remaining trainable pathway. Both counterfactual branches shared all trainable modules; no branch-specific parameters were introduced. The adapted embedding served simultaneously as the input to FiLM, the classifier, feature consistency, and prototype regularization.

The server did not independently average the low-rank factors. Each client induced a full low-rank weight update; the server combined those updates using the applicable aggregation weights; and the resulting matrix was refactorized to the configured rank via a compact QR/SVD procedure. This delta-SVD approach was applied to FedHisto-PAST and the architecture-matched PEFT comparators, preventing the comparison from being confounded by different LoRA aggregation rules.

For a targeted weight matrix, the client-induced and server-combined updates were defined as follows.

\begin{equation}
\begin{aligned}
\Delta W_k &= \frac{\alpha}{r}B_kA_k,\\
\Delta W_{\mathrm{server}}
&=
\sum_k a_k\Delta W_k.
\end{aligned}
\label{eq:lora-server-update}
\end{equation}

The final method processed both stain-counterfactual views through the same frozen encoder, LoRA layers, residual adapter, stain-conditioned FiLM pathway, and classifier. The local objective combined supervised classification with prediction, feature, and prototype regularization.

\begin{equation}
\mathcal{L}
=
\mathcal{L}_{\mathrm{cls}}
+
0.25\mathcal{L}_{\mathrm{JS}}
+
0.10\mathcal{L}_{\mathrm{feat}}
+
0.05\mathcal{L}_{\mathrm{proto}}.
\label{eq:local-objective}
\end{equation}

The classification term was effective-number class-balanced cross-entropy \cite{ref06}, applied symmetrically to paired predictions. The effective-number parameter was $\beta=0.999$; weights were normalized per available client classes and capped at 4.0. A locally absent class received zero weight, and replacement oversampling was not used. Consequently, C5 received no synthetic SCC supervision. This class-loss formulation was held constant across the architecture-matched principal methods so that the proposed objectives were not compared against a different imbalance strategy.

\subsection{Consistency Losses and Reliability-Aware Prototype Regularization}
\label{subsec:consistency-prototypes}

Prediction consistency was defined using the Jensen-Shannon divergence between the categorical distributions produced by views A and B. Supervised losses were applied symmetrically, and neither view acted as a fixed teacher. The implementation formed the mixture distribution with floating-point precision and bounded its probabilities away from zero to ensure numerical stability. Feature consistency operated on the paired adapted embeddings before classification: both embeddings were L2-normalized, and their mean cosine distance was minimized. These terms enforced agreement at complementary output and representation levels while retaining the shared tissue geometry established in Section~\ref{subsec:image-preprocessing}.

Prototype learning followed the principle of federated class-level representation exchange \cite{ref14}, while explicitly respecting class availability. After model aggregation, each client computed normalized class centroids from a deterministic subset of at most 256 optimization examples per locally observed class. A missing class generated neither a placeholder nor a zero prototype; therefore, C5 submitted ACA and Normal prototypes but never an SCC prototype. The server aggregated prototypes only where availability masks were true, using class counts raised to the power 0.5. The server object retained global class centroids, contribution counts, and availability masks separately from the model state.

Prototype reliability was derived from server-side variance when such evidence was available. Per-class variance was reduced across feature dimensions, floored at $10^{-4}$, and inverted. When variance was unavailable, reliability used the count-based fallback $\sqrt{n}/(\sqrt{n}+5)$. Reliability was multiplied by the class-availability mask, normalized to have a mean of 1 across valid classes, and clipped to $[0,3]$. Samples whose labels lacked a valid global prototype were omitted from this term. The regularizer combined reliability-weighted cosine attraction to the correct prototype with a hardest-available-negative margin of 0.20.

\begin{equation}
\begin{aligned}
\mathcal{L}_{\mathrm{proto}}
&=
\operatorname{mean}_i
\left\{
r(y_i)
\left[
1-s\!\left(z_i,p(y_i)\right)
\right]
\right\}
\\
&\quad+
\operatorname{mean}_i
\max
\left\{
0,\,
0.20
+
\max_{c\neq y_i}
s\!\left(z_i,p(c)\right)
-
s\!\left(z_i,p(y_i)\right)
\right\}.
\end{aligned}
\label{eq:prototype-regularizer}
\end{equation}

The component weights of 0.25, 0.10, and 0.05 and the prototype margin of 0.20 were fixed before the reported exploratory evaluation. Three component-removal configurations omitted prediction consistency, feature consistency, or prototype regularization while retaining the remaining architecture, optimization, aggregation, and data protocol. These experiments were designed to assess dependence on individual terms within the frozen configuration, not to assume that every component contributed independently.

\subsection{Federated Optimization, Aggregation, and Comparators}
\label{subsec:federated-optimization}

At each communication round, the server distributed the current state of the trainable model and, where applicable, the current global prototypes. Every participating client initialized from that shared state, performed two local epochs, and returned its updated trainable state together with prototypes for locally available classes. The canonical complete rounds retained all five planned clients; the implementation aggregated the successfully returned states rather than fabricating an update for a missing return. Model updates and class prototypes were serialized and aggregated as distinct objects. Client optimization used AdamW with learning rate $10^{-4}$, weight decay $10^{-4}$, automatic mixed precision, and gradient clipping at L2 norm 5.0. Paired-view training used a micro-batch of four and accumulated gradients for two steps, giving an effective batch size of eight. Baselines and FedHisto-PAST v1 used a batch size of 8, and all evaluations used a batch size of 32. No learning-rate scheduler was documented in the canonical implementation.

FedAvg-PEFT used sample-mass-weighted aggregation \cite{ref28}. FedProx-PEFT retained the same PEFT capacity and added a proximal penalty with coefficient $\mu=0.001$ relative to the received global state \cite{ref29}. FiLM-Zero added the FiLM network but supplied a constant zero descriptor, used no prototype mechanism, and retained sample-weighted aggregation. FedHisto-PAST v1 combined stain-conditioned FiLM, its availability-aware prototype pathway, and adaptive aggregation without paired-view consistency. The final FedHisto-PAST configuration retained that global structure while introducing paired stain-counterfactual local training, prediction, and feature consistency, and reliability-weighted cosine-margin prototypes. Table~\ref{tab:compact-comparators} summarizes the comparator and ablation design.

\begin{table*}[t]
\centering
\caption{Compact comparator and ablation design}
\label{tab:compact-comparators}
\scriptsize
\setlength{\tabcolsep}{3pt}
\renewcommand{\arraystretch}{1.12}

\begin{tabularx}{\textwidth}{
@{}
>{\raggedright\arraybackslash}p{2.25cm}
>{\raggedright\arraybackslash}p{3.0cm}
>{\raggedright\arraybackslash}p{2.4cm}
>{\raggedright\arraybackslash}p{2.8cm}
>{\raggedright\arraybackslash}p{1.8cm}
Y
@{}
}
\toprule
Method &
Adaptation/conditioning &
Consistency &
Prototype mechanism &
Aggregation &
Executed evidence \\
\midrule

FedAvg-PEFT &
LoRA + adapter + classifier &
None &
None &
Sample-weighted &
Internal 5; external 5 \\

FedProx-PEFT &
FedAvg capacity; proximal loss &
None &
None &
Sample-weighted; $\mu=0.001$ &
Internal 4; external 5 \\

FiLM-Zero &
PEFT + zero-conditioned FiLM &
None &
None &
Sample-weighted &
Internal 5; external 5 \\

FedHisto-PAST v1 &
PEFT + stain FiLM &
Single-view &
Availability-aware v1 &
Adaptive &
Internal 5; external whole-image 5; patch-grid sensitivity secondary \\
FedHisto-PAST v2 &
PEFT + stain FiLM &
Paired JS + feature &
Reliability-aware cosine-margin &
Adaptive &
Internal 5; external whole-image 5 \\
No prediction consistency &
Full method otherwise &
JS removed; feature retained &
Unchanged &
Unchanged &
Internal 3; external 3 (seeds 11/33/55) \\
No feature consistency &
Full method otherwise &
JS retained; feature removed &
Unchanged &
Unchanged &
Internal 3; external 3 (seeds 11/33/55) \\
No prototype regularization &
Full method otherwise &
Paired JS + feature &
Removed &
Unchanged &
Internal 3; external 3 (seeds 11/33/55) \\
\bottomrule
\end{tabularx}

\vspace{2pt}
\begin{minipage}{\textwidth}
\footnotesize
\textit{Note.} Internal counts denote finalized reports. FedProx seed 44 produced a usable checkpoint but no finalized internal report; no internal value was reconstructed or imputed, and the checkpoint remained eligible for frozen external inference.
\end{minipage}
\end{table*}

For FedHisto-PAST v1 and the final method, the adaptive server rule combined sublinear sample mass, agreement between each client update and a leave-one-out reference update, and deviation of the client update norm from the median norm. The initial weights used a sample-count exponent of 0.5; after normalization, a minimum client weight of 0.03 was enforced, and the weights were renormalized. The server reconstructed induced LoRA updates, applied the adaptive weights, refactorized the aggregated update by delta-SVD, and redistributed the updated trainable state and prototypes.

\begin{equation}
\begin{aligned}
\tilde{a}_k
\propto{}&
n_k^{0.5}
\exp
\left\{
-\left[
1-\cos
\left(
\Delta_k,\bar{\Delta}_{-k}
\right)
\right]
\right\}
\\
&\times
\exp
\left\{
-0.5
\left|
\ln
\left(
\frac{\|\Delta_k\|}
{\operatorname{median}_j\|\Delta_j\|}
\right)
\right|
\right\},
\\
a_k
&=
\operatorname{Normalize}
\left(
\max
\left[
\operatorname{Normalize}(\tilde{a}_k),
0.03
\right]
\right),
\\
\theta^{t+1}
&=
\sum_k a_k\theta_k^{t+1}.
\end{aligned}
\label{eq:adaptive-aggregation}
\end{equation}

This adaptive rule was treated as an empirical aggregation heuristic. It was not interpreted as a trust score, Byzantine-robust aggregation, or a mechanism supported by a new convergence proof. Model and prototype aggregation remained distinct: model weights depended on returned update statistics, whereas prototype aggregation depended on class availability, contribution counts, and reliability evidence.

\subsection{Training Control, Checkpointing, and Reproducibility}
\label{subsec:training-control}

The maximum training budget comprised 50 communication rounds with two local epochs per round. Early stopping, monitored by global-validation Macro-F1, required an improvement of at least $10^{-4}$ and stopped after 10 non-improving rounds. The best trainable state and prototype state selected by validation were retained separately from the latest state and restored for evaluation. Atomic checkpoints were written after completed rounds and stored the state required to resume training, including progress, best-state information, validation history, configuration identity, dataset identity, and supported random-number-generator states. Resume logic required agreement among the method, seed, configuration, and canonical data identity.

The principal stochastic seeds were 11, 22, 33, 44, and 55. Python, NumPy, PyTorch CPU, visible CUDA generators, DataLoader generators, and worker initialization were seeded. cuDNN deterministic mode was enabled, and benchmarking was disabled. Torch deterministic algorithms were not globally enforced, so exact bitwise GPU replay was not guaranteed. Independent seed processes were executed on Kaggle NVIDIA T4 GPUs, with paired workers assigned to the available dual-GPU environment when applicable.

Reproducibility controls included the pinned HIBOU-B revision, canonical manifest, deterministic holdout procedure, configuration checks, atomic checkpoints, sample-level prediction exports, and method- and seed-specific reports. Communication accounting included server-to-client downloads of trainable state, client-to-server uploads, and method-specific auxiliary transfers of prototypes or descriptors. The frozen encoder, locally retained optimizer states and gradients, checkpoint files, and routine logs were excluded. Communication was measured in binary mebibytes using the $2^{20}$-byte convention and included both server-to-client and client-to-server transfers across all five clients. Auxiliary prototype traffic was included where used, while the one-time descriptor exchange was tracked separately for transparency. Full environment inventories, checkpoint-field lists, and tensor-level payload derivations are retained for the Supplementary Methods.

\subsection{Internal Evaluation Protocol}
\label{subsec:internal-evaluation}

For each finalized seed, the validation-selected checkpoint was evaluated once on the fixed global test set and on each client-specific holdout. Macro-F1 was the primary endpoint because client and external class distributions were heterogeneous. Additional global metrics included accuracy, balanced accuracy, class-wise precision, recall, and F1; macro one-versus-rest AUROC, macro AUPRC, negative log-likelihood, multiclass Brier score, and expected calibration error, calculated with 15 equal-width confidence bins. Client-level reporting included the principal classification metrics where class availability permitted, worst-client Macro-F1, and across-client variability. Macro measures were unweighted averages over the three target classes; AUROC and AUPRC used one-versus-rest evaluation. Expected calibration error was weighted by each non-empty bin's absolute accuracy-confidence gap and sample proportion, while absent client classes were retained as an explicit interpretation boundary rather than silently treated as observed labels.

Cross-seed means and sample standard deviations were used only in finalized reports. FedAvg-PEFT, FiLM-Zero, FedHisto-PAST v1, and FedHisto-PAST each contributed five internal reports for seeds 11, 22, 33, 44, and 55. FedProx-PEFT contributed four finalized internal reports: seeds 11, 22, 33, and 55. Its seed-44 run reached the configured stopping condition and produced a usable checkpoint, but the finalized internal prediction and report package were absent after a documented checkpoint-validation failure. The missing internal result was neither reconstructed nor imputed; the checkpoint remained eligible only for frozen external inference. The three-component ablations used seeds 11, 33, and 55.

\subsection{LungHist700 Exploratory Cross-Dataset Evaluation and Statistical Analysis}
\label{subsec:lunghist700-evaluation}

LungHist700 \cite{ref44} was used as a development-influenced exploratory cross-dataset cohort. The canonical evaluation manifest contained 691 images: 280 ACA, 151 Normal, and 260 SCC. For each frozen validation-selected checkpoint, images underwent a deterministic whole-image resize to $224 \times 224$, canonical normalization, and a single forward pass per image. FedHisto-PAST used the fixed centroid of the training-client stain descriptors. No LungHist700 image was used for local optimization, server aggregation, early stopping, checkpoint selection, probability threshold adjustment, post hoc calibration, or parameter modification after the evaluated configuration was frozen.

Five checkpoint-specific probability vectors were aligned by stable image identifier and averaged image-by-image; the ensemble label was the argmax of the averaged three-class probability. Thus, each method-level ensemble used $S=5$ frozen checkpoints, including the available FedProx seed-44 checkpoint, without changing the internal-report inclusion rule described in Section~\ref{subsec:internal-evaluation}. The primary comparison used the same deterministic whole-image protocol for the eligible methods. The secondary patch-grid sensitivity analysis for FedHisto-PAST v1 and the complete subgroup definitions were retained outside the compact main Methodology. Available magnification and differentiation metadata were used only for descriptive subgroup summaries and did not affect model selection.

A paired image-level bootstrap analysis compared methods using the same 691 observations. In each of 10,000 replicates, image indices were sampled with replacement, and both methods were evaluated on the identical resampled set. The observed paired difference was retained as the point estimate. Percentile 95\% confidence intervals were obtained from the 2.5th and 97.5th percentiles of the bootstrap distribution, and unadjusted two-sided paired probabilities or $p$-values were reported where applicable. No multiplicity correction was applied because these pairwise analyses were exploratory. Patient identifiers were unavailable; resampling was therefore performed at the image level rather than patient-clustered. The resulting uncertainty may be understated when images are dependent on one another within a patient or slide; accordingly, the resulting intervals must not be interpreted as patient-level confidence or clinical validation.

\subsection{Privacy, Reproducibility, and Claim Boundaries}
\label{subsec:privacy-boundaries}

Images and labels remained within their assigned simulated-client partitions. The server received trainable model states, class-prototype summaries, and low-dimensional stain descriptors. This raw-data locality reduced direct image centralization but did not constitute a formal privacy guarantee. The study did not implement differential privacy, secure aggregation, homomorphic encryption, a trusted execution environment, a communication-encryption audit, membership-inference testing, gradient-inversion testing, or formal information-leakage analysis. Raw images were local by protocol, but transmitted states and summaries were not evaluated against update-level attack models. The appropriate description is a raw-data-local federated simulation, or a privacy-aware federated design without formal privacy guarantees.

Reproducibility was supported by a pinned model revision, a canonical dataset manifest, seeded execution, deterministic local holdouts, configuration-identity checks, atomic checkpointing, prediction exports, and method- and seed-specific reports. Nevertheless, exact bitwise replay was not guaranteed, and not every execution was cryptographically bound to a preserved source-code snapshot. These controls support the auditability of the reported image-level experiments without establishing complete execution equivalence across future software and hardware environments.

The methodological claims are limited to the five-client image-level simulation that was executed. Unavailable patient and slide identifiers preclude claims of patient-level or slide-level independence. The study did not evaluate operational hospitals, leave-one-client-out generalization, prospective deployment, diagnostic equivalence to pathologists, or clinical readiness. The adaptive aggregation rule provides neither a Byzantine-robustness guarantee nor a new proof of convergence. No formal privacy claim is made, and the exploratory LungHist700 analysis is not presented as independent clinical validation.


\section{Results}
\label{sec:results}

\subsection{Execution Coverage and Evidence Integrity}

The planned internal benchmark used five stochastic seeds (11, 22, 33, 44, and 55) for each of five federated configurations. FedAvg-PEFT, FiLM-Zero, FedHisto-PAST v1, and FedHisto-PAST v2 produced finalized internal reports for all five seeds. FedProx-PEFT produced four finalized internal reports; seed 44 ended with a documented terminal checkpoint-validation failure caused by a mismatch between the recorded history length and the completed-round count. No score was reconstructed or imputed. FedProx-PEFT internal summaries therefore use \(n=4\), whereas the other main methods use \(n=5\). The corresponding seed-44 FedProx checkpoint remained available and was used only for the five-checkpoint LungHist700 ensemble, thereby preserving the distinction between finalized internal reporting and external checkpoint availability.

The fixed internal validation and test partitions contained 1,419 and 1,420 images, respectively, and were held constant across methods and seeds. The development-influenced exploratory LungHist700 cohort contained 691 images. Under the implemented clearance criteria, the evidence bundle reported no detected exact or probable cross-partition family overlap, no global validation-to-test family overlap, and no cross-client training family overlap. These integrity checks supported the controlled image-level comparison but did not establish patient- or slide-level independence.

Across the planned 25 principal internal runs, 24 finalized reports were available. The sole incomplete report was FedProx-PEFT seed 44. Its failure record indicated a RuntimeError during temporary checkpoint validation, where the stored history length was inconsistent with the completed-round count. Because the failure occurred at the reporting and checkpoint-validation stage, its checkpoint could be retained for frozen external inference; however, the absence of final internal predictions and a report excluded it from the internal cross-seed mean. This evidence rule was applied consistently rather than treating checkpoint availability as equivalent to a finalized internal result.

\begin{table*}[t]
\centering
\caption{Execution coverage and internal fixed-split evaluation summary}
\label{tab:execution-coverage}
\scriptsize
\setlength{\tabcolsep}{3pt}
\renewcommand{\arraystretch}{1.15}

\resizebox{\textwidth}{!}{%
\begin{tabular}{
@{}
l
c
c
c
c
c
c
c
c
l
@{}
}
\toprule
Method &
\makecell{Finalized\\$n$} &
Macro-F1 &
Accuracy &
\makecell{Balanced\\accuracy} &
ECE &
Brier &
\makecell{SCC\\recall} &
\makecell{Worst-client\\Macro-F1} &
Execution status \\
\midrule

FedAvg-PEFT &
5 &
\(0.999012 \pm 0.001070\) &
\(0.999014 \pm 0.001068\) &
\(0.999012 \pm 0.001070\) &
\(0.001084 \pm 0.000863\) &
\(0.001817 \pm 0.001790\) &
\(0.998305 \pm 0.002321\) &
\(0.988713 \pm 0.006396\) &
Complete \\

FedProx-PEFT &
5 &
\(0.998589 \pm 0.001577\) &
\(0.998592 \pm 0.001575\) &
\(0.998588 \pm 0.001579\) &
\(0.000999 \pm 0.000816\) &
\(0.001851 \pm 0.001506\) &
\(0.996610 \pm 0.004130\) &
\(0.985585 \pm 0.010183\) &
Complete \\

FiLM-Zero &
5 &
\(0.998449 \pm 0.001684\) &
\(0.998451 \pm 0.001681\) &
\(0.998448 \pm 0.001685\) &
\(0.001316 \pm 0.001244\) &
\(0.002298 \pm 0.002364\) &
\(0.997034 \pm 0.003545\) &
\(0.979749 \pm 0.012749\) &
Complete \\

FedHisto-PAST v1 &
5 &
\(0.998872 \pm 0.000631\) &
\(0.998873 \pm 0.000630\) &
\(0.998872 \pm 0.000630\) &
\(0.001055 \pm 0.000560\) &
\(0.001836 \pm 0.001164\) &
\(0.998729 \pm 0.001160\) &
\(0.994357 \pm 0.007763\) &
Complete \\

FedHisto-PAST v2 &
5 &
\(0.997321 \pm 0.002140\) &
\(0.997324 \pm 0.002136\) &
\(0.997318 \pm 0.002141\) &
\(0.002627 \pm 0.002075\) &
\(0.004843 \pm 0.003653\) &
\(0.994915 \pm 0.005318\) &
\(0.977155 \pm 0.019807\) &
Complete \\

\bottomrule
\end{tabular}%
}
\end{table*}

\noindent\textit{Note.} Values are mean \(\pm\) sample standard deviation across finalized internal seeds. FedProx-PEFT uses \(n=4\) because seed 44 did not yield a complete finalized internal report; all other methods use \(n=5\). No failed score was imputed.

\subsection{Internal Fixed-Split Performance}

All methods reached near-ceiling performance on the fixed global test set (Table~\ref{tab:execution-coverage}). FedAvg-PEFT achieved the highest five-seed mean Macro-F1 among methods with complete five-seed internal evidence (\(0.999012 \pm 0.001070\)). FedProx-PEFT had a slightly higher mean of \(0.999295 \pm 0.000000\); however, that estimate was based on four finalized reports and is not a directly matched five-seed summary. FedHisto-PAST v1 achieved \(0.998872 \pm 0.000631\), FiLM-Zero achieved \(0.998449 \pm 0.001684\), and FedHisto-PAST v2 achieved \(0.997321 \pm 0.002140\). The corresponding accuracy and balanced accuracy for v2 were \(0.997324 \pm 0.002136\) and \(0.997318 \pm 0.002141\), respectively.

Internal calibration errors were also small in absolute terms. FedHisto-PAST v2 had an ECE of \(0.002627 \pm 0.002075\) and a Brier score of \(0.004843 \pm 0.003653\). Its SCC recall was \(0.994915 \pm 0.005318\). These values confirmed high in-domain separability but did not establish the internal superiority of v2. Because all configurations achieved near-ceiling classification, the small internal rank differences provided limited discrimination among the methods and were not used as the principal basis for the performance claim.

The same pattern was evident across the remaining internal metrics. FedAvg-PEFT produced balanced accuracy \(0.999012 \pm 0.001070\), ECE \(0.001084 \pm 0.000863\), Brier score \(0.001817 \pm 0.001790\), and SCC recall \(0.998305 \pm 0.002321\). The four finalized FedProx-PEFT reports yielded balanced accuracy \(0.999294 \pm 0.000001\), ECE \(0.000657 \pm 0.000331\), Brier score \(0.001194 \pm 0.000374\), and SCC recall \(0.998411 \pm 0.001059\). FiLM-Zero and FedHisto-PAST v1 achieved balanced accuracies of \(0.998448 \pm 0.001685\) and \(0.998872 \pm 0.000630\), respectively. Their ECE values were \(0.001316 \pm 0.001244\) and \(0.001055 \pm 0.000560\). These differences were numerically small relative to the common near-ceiling level and did not provide a reliable basis for ordering the methods.

\subsection{Client-Level Robustness}

Worst-client Macro-F1 was examined to determine whether high global-test scores concealed weaker source-local behaviour. FedHisto-PAST v1 had the highest mean worst-client Macro-F1 (\(0.994357 \pm 0.007763\)), followed by FedProx-PEFT (\(0.989156 \pm 0.007297\)) and FedAvg-PEFT (\(0.988713 \pm 0.006396\)). FiLM-Zero achieved \(0.979749 \pm 0.012749\), while FedHisto-PAST v2 achieved \(0.977155 \pm 0.019807\). The v2 value showed the largest between-seed dispersion among the reported methods. Across its five seeds, the reported worst-client Macro-F1 ranged from 0.946208 to 0.998584. Overall, the proposed configuration retained high client-level performance but did not improve the worst-client internal summary compared with the architecture-matched comparators.

These client-level results are reported separately from the global average because the five simulated clients had unequal sample sizes and class coverage. The analysis does not identify a single client as uniformly worst across seeds; it summarizes the minimum client-holdout performance observed within each completed run.

The per-seed worst-client ranges further illustrate the difference in stability. FedAvg-PEFT ranged from 0.984841 to 1.000000, FedProx-PEFT from 0.984841 to 1.000000 across its four finalized reports, FiLM-Zero from 0.971301 to 1.000000, and FedHisto-PAST v1 from 0.984841 to 1.000000. FedHisto-PAST v2 ranged from 0.946208 to 0.998584. The lower v2 endpoint occurred despite a global-test Macro-F1 above 0.995 for that seed, showing why global performance and worst-client performance were retained as distinct evidence dimensions.

\subsection{Exploratory LungHist700 Cross-Dataset Performance}

LungHist700 comprised 691 images: 280 ACA, 151 Normal, and 260 SCC. The cohort was development-influenced and therefore treated as an exploratory cross-dataset evaluation rather than an untouched external validation cohort. After the evaluated FedHisto-PAST v2 configuration was frozen, no further fine-tuning, threshold adjustments, post hoc calibration, or parameter modifications were performed. Five checkpoint-specific probability vectors were averaged image by image to form each method-level ensemble.

\begin{table*}[t]
\centering
\caption{LungHist700 whole-image probability-ensemble performance}
\label{tab:lunghist700}
\scriptsize
\setlength{\tabcolsep}{3pt}
\renewcommand{\arraystretch}{1.15}

\resizebox{\textwidth}{!}{%
\begin{tabular}{
@{}
l
c
c
c
c
c
c
c
c
c
c
c
@{}
}
\toprule
Method &
Accuracy &
\makecell{Balanced\\accuracy} &
Macro-F1 &
Macro-AUROC &
Macro-AUPRC &
NLL &
Brier &
ECE &
\makecell{ACA\\recall} &
\makecell{Normal\\recall} &
\makecell{SCC\\recall} \\
\midrule

FedAvg-PEFT &
0.677279 &
0.678285 &
0.674942 &
0.902991 &
0.841937 &
4.074109 &
0.590220 &
0.273546 &
0.989286 &
0.741722 &
0.303846 \\

FedProx-PEFT &
0.693198 &
0.691828 &
0.696334 &
0.923432 &
0.886197 &
2.173750 &
0.541572 &
0.251508 &
0.975000 &
0.735099 &
0.365385 \\

FiLM-Zero &
0.674385 &
0.672019 &
0.672025 &
0.901589 &
0.847102 &
3.610070 &
0.593431 &
0.282894 &
0.989286 &
0.715232 &
0.311538 \\

FedHisto-PAST v1 &
0.677279 &
0.681427 &
0.677191 &
0.875529 &
0.806098 &
5.487567 &
0.587681 &
0.287265 &
0.975000 &
0.761589 &
0.307692 \\

FedHisto-PAST v2 &
0.717800 &
0.730454 &
0.728560 &
0.913336 &
0.848381 &
3.648615 &
0.513336 &
0.248052 &
0.932143 &
0.847682 &
0.411538 \\

\bottomrule
\end{tabular}%
}
\end{table*}

\noindent\textit{Note.} All values are five-checkpoint probability-ensemble results on the same 691 images under the primary deterministic whole-image protocol. The separate FedHisto-PAST v1 patch-grid experiment was retained only as a secondary sensitivity analysis.

FedHisto-PAST v2 achieved an ensemble accuracy of 0.717800, balanced accuracy of 0.730454, and Macro-F1 of 0.728560 (Table~\ref{tab:lunghist700}). Its Macro-AUROC and Macro-AUPRC were 0.913336 and 0.848381. The corresponding NLL, Brier score, and ECE were 3.648615, 0.513336, and 0.248052. Comparator Macro-F1 values under the same whole-image ensemble protocol were 0.674942 for FedAvg-PEFT, 0.696334 for FedProx-PEFT, 0.672025 for FiLM-Zero, and 0.677191 for FedHisto-PAST v1.

Across checkpoint-specific evaluations, v2 had a mean Macro-F1 of \(0.729723 \pm 0.022381\), compared with \(0.666073 \pm 0.046689\) for FedAvg-PEFT, \(0.690303 \pm 0.032167\) for FedProx-PEFT, \(0.667589 \pm 0.036594\) for FiLM-Zero, and \(0.683135 \pm 0.029312\) for FedHisto-PAST v1. The five v2 checkpoint-specific Macro-F1 values ranged from 0.703047 to 0.762078. Therefore, the probability ensemble remained close to the center of the seed-level distribution rather than being driven by a single isolated checkpoint.

For v2, ACA precision, recall, and F1 were 0.598624, 0.932143, and 0.729050; Normal precision, recall, and F1 were 0.941176, 0.847682, and 0.891986; and SCC precision, recall, and F1 were 0.899160, 0.411538, and 0.564644. Relative to the eligible whole-image comparators, v2 produced higher Normal and SCC recall but lower ACA recall. It therefore improved cross-class balance rather than dominating every class.

The v2 confusion matrix contained 261 correct ACA predictions, eight ACA images predicted as Normal, and 11 predicted as SCC. Among Normal images, 128 were correctly classified, 22 were predicted as ACA, and one was predicted as SCC. For SCC, 107 images were correctly classified, and 153 were predicted as ACA. In comparison, SCC recall was 0.303846 for FedAvg-PEFT, 0.365385 for FedProx-PEFT, 0.311538 for FiLM-Zero, and 0.307692 for FedHisto-PAST v1, while the corresponding ACA recalls were 0.989286, 0.975000, 0.989286, and 0.975000. This class redistribution explains why v2 improved Macro-F1 and balanced accuracy despite a lower ACA recall.

FedHisto-PAST v1 was evaluated using the primary deterministic whole-image protocol and achieved an ensemble Macro-F1 of 0.677191. A separate deterministic patch-grid sensitivity analysis produced an ensemble Macro-F1 of 0.539225 and was retained only as secondary evidence because it combines the v1 model with a different inference procedure.



\begin{figure*}[t]
\centering
\includegraphics[width=\textwidth]{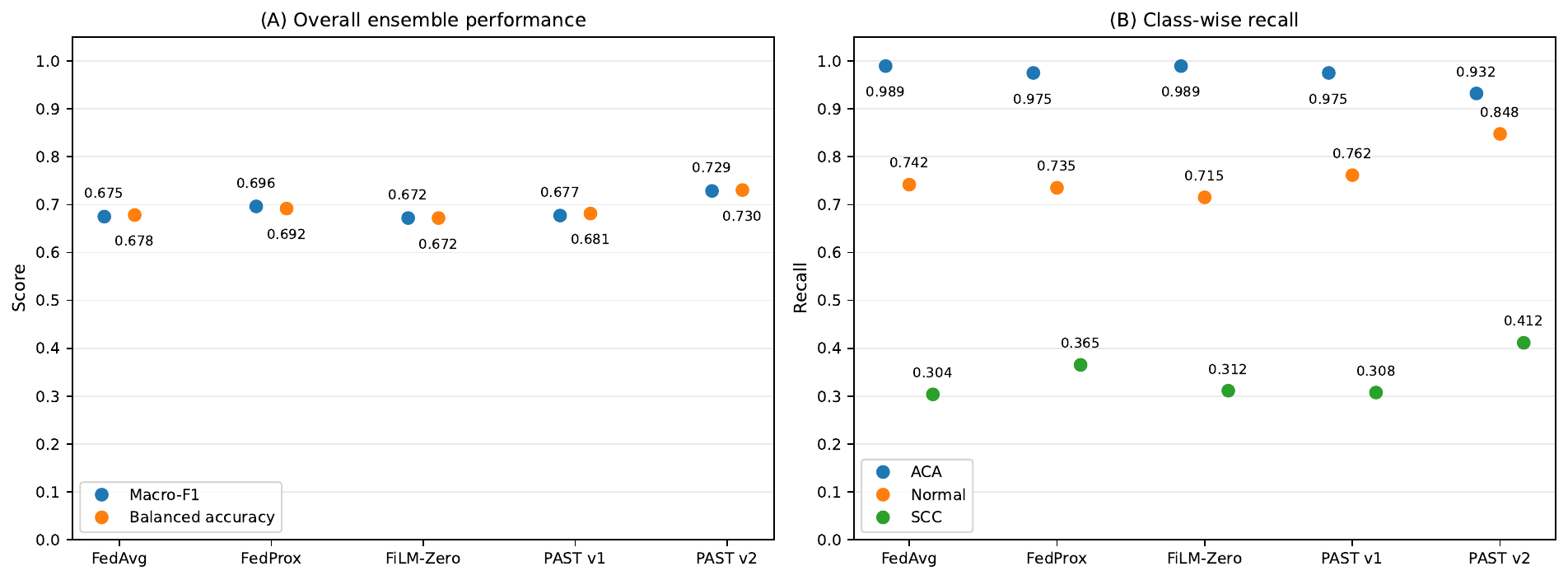}
\caption{Exploratory LungHist700 whole-image ensemble performance.
(A) Macro-F1 and balanced accuracy across the five evaluated
federated configurations. (B) Class-wise recall for adenocarcinoma
(ACA), Normal, and squamous cell carcinoma (SCC). FedHisto-PAST v2
improved Normal and SCC recall while reducing ACA recall relative
to the whole-image comparators.}
\label{fig:lunghist700-performance}
\end{figure*}


\subsection{Paired Image-Level Bootstrap Comparisons}

The ensemble predictions were aligned image by image across the 691 LungHist700 observations. The paired analysis used 10,000 bootstrap resamples with replacement and RNG seed 20260730. Because usable patient identifiers were unavailable, images were used as the resampling unit, and the resulting intervals are not patient-clustered clinical confidence intervals. No multiplicity adjustment was applied to these exploratory pairwise analyses.

\begin{table}[t]
\centering
\caption{Paired image-level LungHist700 bootstrap comparisons}
\label{tab:bootstrap}
\scriptsize
\setlength{\tabcolsep}{2.5pt}
\renewcommand{\arraystretch}{1.18}

\begin{tabularx}{\columnwidth}{
@{}
>{\raggedright\arraybackslash}p{2.15cm}
>{\centering\arraybackslash}p{2.25cm}
>{\centering\arraybackslash}p{1.35cm}
Y
@{}
}
\toprule
Comparator &
\makecell{Macro-F1 difference\\95\% CI} &
\makecell{Two-sided\\bootstrap \(p\)} &
\makecell{Recall difference\\SCC / ACA} \\
\midrule

FedAvg-PEFT &
\makecell{0.053618\\{[0.027328, 0.079784]}} &
\(p < 0.0002\) &
\makecell{0.107692\\\(-0.057143\)} \\

FedProx-PEFT &
\makecell{0.032226\\{[0.008432, 0.056709]}} &
\(p = 0.0090\) &
\makecell{0.046154\\\(-0.042857\)} \\

FiLM-Zero &
\makecell{0.056535\\{[0.029800, 0.083942]}} &
\(p < 0.0002\) &
\makecell{0.100000\\\(-0.057143\)} \\

FedHisto-PAST v1
(secondary patch-grid pipeline) &
\makecell{0.189335\\{[0.149462, 0.231351]}} &
\(p < 0.0002\) &
\makecell{0.184615\\\(-0.039286\)} \\

\bottomrule
\end{tabularx}
\end{table}

\noindent\textit{Note.} Differences are FedHisto-PAST v2 minus comparator. Positive values favour v2. All rows use the same primary deterministic whole-image protocol. No multiplicity adjustment was applied. The separate v1 patch-grid comparison was excluded from this primary table.

Compared with FedAvg-PEFT, v2 increased Macro-F1 by 0.053618 (95\% CI [0.027328, 0.079784]; \(p < 0.0002\)) and SCC recall by 0.107692, while ACA recall decreased by 0.057143. Relative to FedProx-PEFT, the Macro-F1 difference was 0.032226 (95\% CI [0.008432, 0.056709]; \(p = 0.0090\)), SCC recall increased by 0.046154, and ACA recall decreased by 0.042857. The corresponding Macro-F1 difference versus FiLM-Zero was 0.056535 (95\% CI [0.029800, 0.083942]; \(p < 0.0002\)), with an SCC-recall increase of 0.100000 and an ACA-recall decrease of 0.057143.

The secondary comparison with the v1 patch-grid pipeline yielded a Macro-F1 difference of 0.189335 (95\% CI [0.149462, 0.231351]; \(p < 0.0002\)), an SCC-recall difference of 0.184615, and an ACA-recall difference of \(-0.039286\). Across the three primary paired comparisons, the image-level intervals supported higher Macro-F1 and SCC sensitivity for v2, while ACA recall was lower. Because the analyses were exploratory and unadjusted for multiplicity, the \(p\)-values are reported descriptively rather than as confirmatory evidence. The separate FedHisto-PAST v1 patch-grid comparison was retained only as a secondary sensitivity analysis and was not included in the primary paired table.

The broader paired metric panel was directionally consistent with the Macro-F1 findings. Versus FedAvg-PEFT, v2 increased balanced accuracy by 0.052170 (95\% CI [0.027232, 0.076809]), accuracy by 0.040521 (95\% CI [0.015919, 0.063676]), and Normal recall by 0.105960 (95\% CI [0.054422, 0.161491]). Versus FedProx-PEFT, the corresponding differences were 0.038626 for balanced accuracy, 0.024602 for accuracy, and 0.112583 for Normal recall. Versus FiLM-Zero, v2 increased balanced accuracy by 0.058436, accuracy by 0.043415, and Normal recall by 0.132450. The secondary v1 pipeline comparison produced differences of 0.185309 in balanced accuracy, 0.143271 in accuracy, and 0.410596 in Normal recall. Each comparison used the same resampled image indices for the paired methods.

\subsection{Component Ablation Analysis}

The component ablations used only seeds 11, 33, and 55. The full-v2 reference in this subsection is therefore the matching three-seed subset and must not be confused with the primary five-seed ensemble. The full three-seed v2 ensemble achieved Macro-F1 0.726936, balanced accuracy 0.730281, ACA recall 0.925000, Normal recall 0.854305, and SCC recall 0.411538. External ablation uncertainty was estimated from 5,000 paired image-level bootstrap replicates.









\begin{table*}[t]
\centering
\caption{Three-seed internal Macro-F1 and external ensemble component-ablation results}
\label{tab:ablation}
\scriptsize
\setlength{\tabcolsep}{3pt}
\renewcommand{\arraystretch}{1.15}

\resizebox{\textwidth}{!}{%
\begin{tabular}{
@{}
l
c
c
c
c
c
c
c
c
c
@{}
}
\toprule
Configuration &
\makecell{Internal\\Macro-F1} &
\makecell{External\\Macro-F1} &
\makecell{Balanced\\accuracy} &
\makecell{ACA\\recall} &
\makecell{Normal\\recall} &
\makecell{SCC\\recall} &
\makecell{Full v2\\\(-\) ablation} &
95\% CI &
\makecell{Bootstrap\\\(P(\Delta>0)\)} \\
\midrule

Full v2 (seeds 11/33/55) &
\(0.997885 \pm 0.002444\) &
0.726936 &
0.730281 &
0.925000 &
0.854305 &
0.411538 &
Reference &
--- &
--- \\

Without prediction consistency &
\(0.996944 \pm 0.001077\) &
0.703363 &
0.714530 &
0.939286 &
0.854305 &
0.350000 &
0.023573 &
[0.009206, 0.039099] &
1.0000 \\

Without feature consistency &
\(0.997886 \pm 0.001867\) &
0.732073 &
0.734484 &
0.925000 &
0.847682 &
0.430769 &
\(-0.005138\) &
[\(-0.017560\), 0.006767] &
0.2072 \\

Without prototype regularization &
\(0.997415 \pm 0.001469\) &
0.737960 &
0.736469 &
0.939286 &
0.827815 &
0.442308 &
\(-0.011024\) &
[\(-0.025819\), 0.002940] &
0.0614 \\

\bottomrule
\end{tabular}%
}

\vspace{2pt}
\begin{minipage}{\textwidth}
\footnotesize
\textit{Note.} All configurations used seeds 11, 33, and 55.
Internal Macro-F1 values are reported as mean \(\pm\) sample standard
deviation across the three completed seeds. External values represent
three-checkpoint probability ensembles evaluated on the same 691
LungHist700 images. Differences are defined as
\(\Delta=\) full-v2 ensemble Macro-F1 minus ablation ensemble Macro-F1;
therefore, positive values favour the full v2 configuration. The 95\%
confidence intervals and bootstrap proportions \(P(\Delta>0)\) were
estimated from 5,000 paired image-level bootstrap replicates. No
multiplicity adjustment was applied.
\end{minipage}

\end{table*}

Removing prediction-level consistency reduced the external ensemble Macro-F1 from 0.726936 to 0.703363. The paired difference was 0.023573 with 95\% CI [0.009206, 0.039099], and the probability of a positive difference was 1.0000. This was the only removed component for which the reported interval excluded zero in favour of the complete configuration.

Removing feature consistency produced an ensemble Macro-F1 of 0.732073. The full-v2 minus ablation difference was \(-0.005138\) with 95\% CI [\(-0.017560\), 0.006767]. Removing prototype regularization produced an ensemble Macro-F1 of 0.737960 and a difference of \(-0.011024\) with 95\% CI [\(-0.025819\), 0.002940]. Both intervals crossed zero. Therefore, the performed ablations did not establish a conclusive independent mean Macro-F1 benefit for feature consistency or prototype regularization, although the removals altered class-specific trade-offs.

The internal three-seed ablation summaries were also similar. Full v2 achieved internal Macro-F1 \(0.997885 \pm 0.002444\), ECE \(0.001914 \pm 0.002042\), Brier score \(0.003538 \pm 0.003678\), worst-client Macro-F1 \(0.984908 \pm 0.013641\), and SCC recall \(0.996469 \pm 0.004410\). Without prediction consistency, internal Macro-F1 was \(0.996944 \pm 0.001077\) and worst-client Macro-F1 was \(0.980530 \pm 0.007467\). The no-feature and no-prototype variants achieved internal Macro-F1 values of \(0.997886 \pm 0.001867\) and \(0.997415 \pm 0.001469\), respectively. These near-ceiling internal values again provided less separation than the external three-seed ensemble.


\begin{figure*}[t]
\centering
\includegraphics[width=\textwidth]{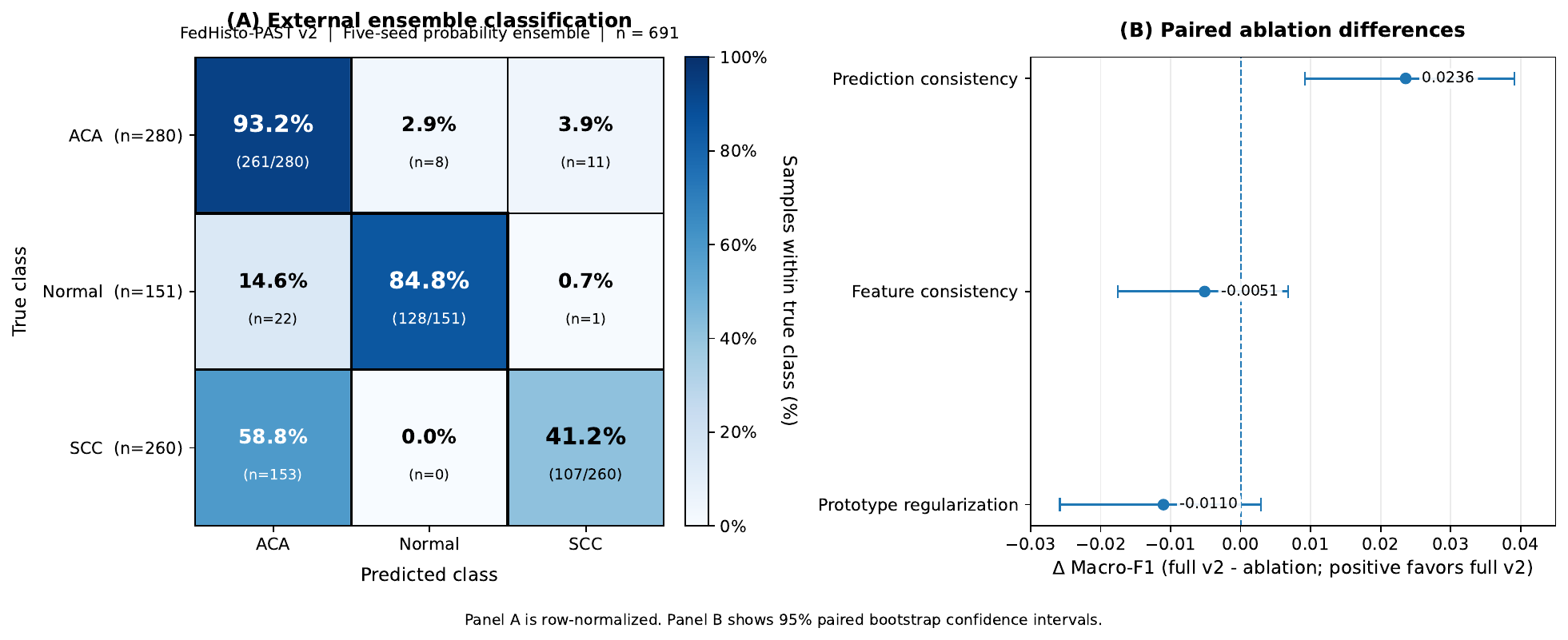}
\caption{Error and component-ablation analysis on LungHist700.
(A) FedHisto-PAST v2 confusion matrix, with rows denoting true
classes and columns denoting predicted classes. The dominant error
was SCC misclassification as ACA. (B) Paired external Macro-F1
differences between full FedHisto-PAST v2 and each component
ablation. Points denote full-v2 minus ablation differences, and
horizontal lines show 95\% image-level bootstrap confidence
intervals from 5,000 replicates.}
\label{fig:error-ablation-analysis}
\end{figure*}

\subsection{Parameter and Communication Efficiency}

FedHisto-PAST v2 updated 1,088,707 of 86,829,763 parameters, corresponding to a trainable fraction of 1.253841\% and a trainable payload of 4.153088 MiB. The per-round communication cost was 41.615891 MiB across five clients. FedAvg-PEFT and FedProx-PEFT each communicated 37.685661 MiB per round, resulting in a v2 overhead of 10.428978\%. The one-time stain-descriptor upload was approximately 0.000267 MiB. These quantities describe only parameter and communication accounting; reductions in runtime, energy use, memory use, and floating-point operations were not measured.

For the architecture-matched FedAvg-PEFT and FedProx-PEFT baselines, 987,907 parameters were trainable, representing 1.137752\% of the complete model and a 3.768566 MiB trainable payload. FiLM-Zero communicated 41.530876 MiB per round, an overhead of 10.203390\% relative to FedAvg-PEFT. FedHisto-PAST v1 and v2 each communicated 41.615891 MiB per round. Canonical trainable-parameter artifacts were unavailable for FiLM-Zero or v1, and those cells were left unreported rather than inferred.

\begin{table*}[t]
\centering
\caption{Parameter and communication efficiency}
\label{tab:efficiency}
\scriptsize
\setlength{\tabcolsep}{4pt}
\renewcommand{\arraystretch}{1.15}

\resizebox{\textwidth}{!}{%
\begin{tabular}{
@{}
l
c
c
c
c
c
c
@{}
}
\toprule
Method &
\makecell{Total\\parameters} &
\makecell{Trainable\\parameters} &
\makecell{Trainable\\fraction (\%)} &
\makecell{Trainable\\payload (MiB)} &
\makecell{Communication\\per round (MiB)} &
\makecell{Overhead vs\\FedAvg-PEFT (\%)} \\
\midrule

FedAvg-PEFT &
86,829,763 &
987,907 &
1.137752 &
3.768566 &
37.685661 &
0.000000 \\

FedProx-PEFT &
86,829,763 &
987,907 &
1.137752 &
3.768566 &
37.685661 &
0.000000 \\

FiLM-Zero &
86,829,763 &
1,088,707 &
1.253841 &
4.153088 &
41.530876 &
10.203390 \\

FedHisto-PAST v1 &
86,829,763 &
1,088,707 &
1.253841 &
4.153088 &
41.615891 &
10.428978 \\

FedHisto-PAST v2 &
86,829,763 &
1,088,707 &
1.253841 &
4.153088 &
41.615891 &
10.428978 \\

\bottomrule
\end{tabular}%
}
\end{table*}

\noindent\textit{Note.} MiB uses the \(2^{20}\)-byte convention. Optimizer states that the frozen encoder, checkpoint files, retained local gradients, and routine logs are excluded. ``Not reported'' denotes unsupported parameter-count fields that were not inferred.

\subsection{Calibration, Subgroup Behaviour and Error Patterns}

The v2 external ensemble had an NLL of 3.648615, a Brier score of 0.513336, and a 15-bin ECE of 0.248052. No post hoc calibration was fitted. These values indicate that the external probability scale remained imperfect and do not support its interpretation as a measure of clinical confidence.

The dominant external error pattern was SCC misclassification as ACA. Of 260 SCC images, 153 were classified as ACA, 107 were classified correctly as SCC, and none were classified as Normal. Within the supplied differentiation subgroups, v2 accuracy was 0.414141 for better differentiated SCC, 0.484848 for moderately differentiated SCC, and 0.357895 for poorly differentiated SCC. Therefore, poorly differentiated SCC had the lowest reported v2 accuracy among the three SCC differentiation groups. Overall accuracy was 0.701950 at \(20\times\) and 0.734940 at \(40\times\). These subgroup summaries are descriptive and do not identify a biological cause for the observed differences.

The observed pattern was not uniform across all subclasses. ACA subgroup accuracy was 0.932039 for better differentiated images, 0.977778 for moderately differentiated images, and 0.885057 for poorly differentiated images. For the 151 Normal images, the ensemble recall was 0.847682. These descriptive strata were derived from the supplied metadata and were not used for checkpoint selection or threshold setting. No subgroup-level patient clustering was possible because patient identifiers were unavailable.

Internal performance was near ceiling for every principal method, whereas the development-influenced LungHist700 evaluation provided greater discrimination among them. FedHisto-PAST v2 increased exploratory cross-dataset Macro-F1, Normal recall, and SCC recall relative to the eligible whole-image comparators, while ACA recall decreased. Prediction-level consistency had the clearest independently supported ablation contribution; feature consistency and prototype regularization did not show conclusive independent gains in overall Macro-F1. External calibration remained imperfect. The method updated approximately 1.25\% of the model parameters with approximately 10.43\% communication overhead relative to FedAvg-PEFT.


\section{Discussion}
\label{sec:discussion}

\subsection{Principal Findings}
\label{subsec:principal_findings}

The internal benchmark showed that all principal configurations fit the fixed image-level test distribution at near-ceiling performance. Under these conditions, small differences in global Macro-F1 could not establish superiority, and FedHisto-PAST v2 did not lead the internal ranking. Its worst-client summary was also lower and more variable than those of several comparators. The internal results therefore support a narrow conclusion: the proposed configuration remained highly discriminative in-domain, but the fixed split offered limited resolution for distinguishing cross-site performance mechanisms. The development-influenced exploratory cross-dataset evaluation provided the more informative stress test while remaining separate from the internal evidence.

On LungHist700, FedHisto-PAST v2 produced a different class-balance profile from FedAvg-PEFT, FedProx-PEFT, FiLM-Zero, and FedHisto-PAST v1 under the primary whole-image protocol. Higher exploratory Macro-F1 and balanced accuracy were accompanied by improved Normal and SCC recall and reduced ACA recall. The principal result was therefore a redistribution of sensitivity across classes. Most errors involved SCC images classified as ACA, but the available evidence does not support a biological or acquisition-specific explanation. Paired image-level bootstrap comparisons supported positive Macro-F1 and SCC-recall differences against FedAvg-PEFT, FedProx-PEFT, and FiLM-Zero. These analyses were exploratory, unadjusted for multiplicity, and based on images rather than patient-clustered resampling. They provide comparative evidence for the observed image set, without establishing causality or clinical validity.

Other findings further narrow the interpretation. External calibration remained imperfect despite the improved class balance, indicating that better ranking and discrimination did not produce clinically interpretable probability estimates. Among the component removals, prediction-level consistency showed the clearest independently supported contribution, as its removal reduced exploratory external Macro-F1 in an interval excluding zero. Removing feature consistency or prototype regularization did not yield a conclusive independent overall Macro-F1 loss, although both altered class-specific behavior. These results support the paired-prediction objective more strongly than the claim that every regularizer contributes independently. Efficiency accounting showed that the framework updated only a small fraction of the frozen HIBOU-B pathway while incurring modest communication overhead compared with architecture-matched FedAvg-PEFT. The evidence supports parameter-efficient adaptation with explicit cost accounting, not claims of reduced runtime, energy use, or memory consumption.

\subsection{Interpretation in Relation to Prior Work}
\label{subsec:interpretation_prior_work}

The internal ceiling effect is consistent with a broader concern in federated histopathology: aggregate performance on a controlled source can conceal feature shifts, client imbalances, and weak transferability \cite{ref04}. FedAvg and FedProx address distributed optimization and local drift \cite{ref28,ref29}. FedBN and SiloBN localize normalization statistics to reduce feature-distribution mismatch \cite{ref30,ref05}. FedHisto-PAST v2 addresses a related problem at a different level by coupling a shared frozen pathology encoder with client stain descriptors and matched stain-counterfactual views. The observed change in external class balance suggests that local appearance perturbations influenced the learned decision boundary, although the experiment does not isolate stain conditioning as the sole causal factor.

This positioning also differs from federated stain-normalization and distribution-alignment methods. Orchestral normalization and BottleGAN use generative approaches to harmonize client staining \cite{ref07,ref08}. FedSDA and FedStain align stain distributions or higher-order stain statistics \cite{ref09,ref10}. FedHisto-PAST v2 instead penalizes disagreement between two controlled stain realizations of the same tissue at the prediction and representation levels. The ablation result for prediction consistency directly supports this sample-correspondence mechanism within the tested configuration. It does not show that paired consistency is universally preferable to normalization or distribution alignment, since the studies use different tasks, cohorts, and evaluation units.

The foundation-model results are also bounded. HIBOU supplies a reusable pathology representation \cite{ref11}. Independent work shows that the pathology foundation model features can preserve site-linked signatures and that frozen feature extractors vary across downstream tasks and aggregation protocols \cite{ref12,ref40}. Low-rank adaptation provides a practical way to update and communicate a compact task-specific state \cite{ref13}. In the present framework, this efficiency was combined with stain-aware training rather than treated as a robustness mechanism in its own right. Prototype exchange offers a complementary semantic channel across heterogeneous clients \cite{ref14}. The missing-class-aware implementation was methodologically appropriate because C5 contained no SCC examples; however, the ablation evidence did not demonstrate a conclusive independent overall gain from prototype regularization. Its current role is therefore best understood as a structured, availability-aware constraint within the integrated design.

\subsection{Implications for Federated Computational Pathology}
\label{subsec:implications_federated_pathology}

These findings have three implications for experimental practice. First, near-perfect internal accuracy should not be treated as sufficient evidence of federated robustness. Fixed global testing should be complemented by client-level summaries, class-specific sensitivity, calibration, and an evaluation that changes the source distribution. Second, average performance can hide clinically relevant trade-offs. The improvement in exploratory SCC and Normal recognition occurred alongside a reduction in ACA recall, so model selection should consider the intended error costs rather than optimizing a single aggregate metric in isolation. Third, communication-efficient adaptation should be reported quantitatively. Transmitting lightweight trainable modules can make foundation-model federation more tractable, but the additional conditioning and prototype pathways still impose measurable overhead.

The framework also illustrates an important claim boundary. Keeping images within simulated client partitions is raw-data-local and may reduce direct data movement, but it is not equivalent to differential privacy, secure aggregation, or resistance to inference attacks \cite{ref03}. Consequently, the present study supports a privacy-aware experimental design without formal privacy guarantees. Likewise, the LungHist700 findings indicate exploratory cross-dataset robustness, not patient-level independence or prospective clinical validity.

The results motivate federated pathology systems that connect representation adaptation, stain variation, missing-class handling, and evaluation design. FedHisto-PAST v2 provides one auditable implementation of this integration. The evidence shows that a frozen pathology foundation model can be adapted using a compact communicated state and that paired prediction consistency improved cross-dataset class balance in the tested setting. Independent, patient-grouped multicentre evaluation, explicit probability calibration, and formal privacy mechanisms remain necessary before stronger translational claims can be considered.
\subsection{Limitations}
\label{subsec:limitations}

This study used five simulated dataset partitions as federated clients rather than operational hospitals. The unit of analysis was the image because patient and whole-slide identifiers were unavailable. Internal separation was therefore established through path disjointness, exact-file hashing, and the implemented perceptual- and local-feature-family audits, rather than through patient- or slide-grouped partitioning. The absence of detected overlap under these procedures should be interpreted as an image-corpus quality-control result, not as proof of patient-level independence.

LungHist700 was development-influenced and was used only as an exploratory cross-dataset cohort, not as an untouched external clinical validation cohort. The paired bootstrap intervals were calculated at the image level rather than using patient-clustered resampling and may understate uncertainty if multiple images were dependent within the same patient or slide.

The internal benchmark produced near-ceiling results for all principal methods and therefore offered limited discrimination among them. FedProx-PEFT had four finalized internal seed reports because seed 44 ended in a documented report and checkpoint-validation failure; no value was reconstructed or imputed. The available FedHisto-PAST v1 patch-grid evidence may also combine differences in the method with differences in the inference pipeline; therefore, it should not be interpreted as an isolated architectural comparison.

In the exploratory cross-dataset cohort, probability calibration remained weak, and no post hoc calibration was fitted, preventing a clinical-confidence interpretation of the reported probabilities. Improved cross-class balance and SCC sensitivity were accompanied by lower ACA recall, so the method did not improve every class. In addition, the component ablations did not yield conclusive, independent external Macro-F1 gains for feature consistency or prototype regularization, although their removal altered class-specific behavior.

The federated simulation kept source images within their assigned client partitions but did not implement differential privacy, secure aggregation, membership inference testing, gradient inversion testing, or any formal leakage guarantee. The study also did not evaluate prospective deployment, comparison with pathologists, tumour grading, prognosis, or transfer to broader pathology tasks.

Reproducibility controls included fixed seeds, pinned model resources, deterministic partition procedures, and checkpoint records; however, exact bitwise replay was not guaranteed, and not every execution was cryptographically bound to a preserved source-code snapshot. These constraints limit operational reproducibility and the strength of translational conclusions. Stronger evidence will require patient- and slide-grouped multicentre evaluation, calibrated probability assessment, independently preserved execution provenance, and formal privacy testing under a prospectively defined protocol.


\section{Conclusion}
\label{sec:conclusion}

Federated lung histopathology classification must account for stain variation, non-IID class distributions, missing client classes, and the cost of adapting large pathology encoders within the same training process. FedHisto-PAST v2 was developed for three-class classification of adenocarcinoma, Normal, and squamous cell carcinoma in a five-client raw-data-local federated simulation.

The framework combines a frozen HIBOU-B encoder with parameter-efficient, trainable modules; stain-conditioned paired counterfactual views; prediction- and feature-level consistency; class-availability- and reliability-aware prototype regularization; and adaptive model aggregation. The design treats stain variation, lightweight adaptation, and heterogeneous class evidence as one federated learning problem. It does not claim formal privacy protection.

The fixed internal test set yielded near-ceiling performance for all principal configurations and therefore provided limited discrimination among methods. FedHisto-PAST v2 did not establish internal superiority or an improvement in worst-client performance. The development-influenced exploratory LungHist700 evaluation was more discriminative. Relative to the available federated parameter-efficient comparators, FedHisto-PAST v2 improved exploratory cross-dataset Macro-F1, balanced class recognition, Normal recall, and SCC sensitivity. These gains were accompanied by reduced ACA recall, and the external probability estimates remained imperfectly calibrated.

Component-removal analysis identified prediction-level consistency as the only element with a clearly supported independent external contribution. Feature consistency and prototype regularization did not show conclusive independent overall Macro-F1 gains, although they remained part of the integrated formulation and affected class-specific behaviour.

FedHisto-PAST v2 updated approximately 1.25\% of the model's parameters and communicated a compact, trainable state while leaving the foundation model backbone frozen. It incurred a modest but measurable overhead relative to FedAvg-PEFT. The evidence supports a bounded conclusion: stain-aware, parameter-efficient federation with paired prediction consistency improved exploratory cross-dataset class balance and SCC sensitivity under the tested image-level protocol.

These results do not establish universal superiority, clinical readiness, patient-level independence, or formal privacy. Broader clinical claims require patient-grouped, multi-institutional, and prospectively defined validation, together with explicit probability calibration and formal privacy evaluation where such guarantees are intended.




\section*{CRediT authorship contribution statement}

\textbf{Muhammad Muhtasim Shahriar:}
Conceptualization, Methodology, Software, Data curation, Investigation,
Formal analysis, Validation, Visualization, Writing -- original draft.

\textbf{M.\,M. Golam Hafiz:}
Data curation, Investigation, Formal analysis, Validation, Visualization,
Writing -- original draft.

\textbf{Saad Aloteibi:}
 Supervision, Project administration, Methodology, Resources, Funding acquisition, Writing -- review \& editing.

\textbf{Mohammad Ali Moni:}
Supervision, Project administration, Methodology, Resources,
Writing -- review \& editing.

\section*{Declaration of Competing Interest}

The authors declare that they have no known competing financial
interests or personal relationships that could have appeared to
influence the work reported in this paper.



\section*{Data Availability}

The datasets used in this study are publicly available from their
original sources. The internal experimental corpus was constructed
from publicly available lung histopathology images associated with
LC25000 \cite{ref42} and WSSS4LUAD \cite{ref43}, while LungHist700
\cite{ref44} was used for exploratory cross-dataset evaluation.
These datasets are available through the repositories associated with
their original publications. No new clinical data were collected for
this study.






\end{document}